\def\year{2022}\relax
\documentclass[letterpaper]{article} 
\usepackage{aaai22}  
\usepackage{times}  
\usepackage{helvet}  
\usepackage{courier}  
\usepackage[hyphens]{url}  
\usepackage{graphicx} 
\usepackage{natbib}  
\usepackage{caption} 
\DeclareCaptionStyle{ruled}{labelfont=normalfont,labelsep=colon,strut=off} 
\usepackage{algorithm}
\usepackage{algorithmic}

\usepackage{microtype}
\usepackage{multirow}
\usepackage{makecell}
\usepackage{booktabs}
\usepackage{amsmath}
\usepackage{subfigure}
\usepackage{xcolor}

\usepackage{newfloat}
\usepackage{listings}
\floatstyle{ruled}
\newfloat{listing}{tb}{lst}{}
\floatname{listing}{Listing}
\title{Incorporating Constituent Syntax for Coreference Resolution}
\author {
    Fan Jiang and Trevor Cohn
}
\affiliations {
    School of Computing and Information Systems\\
    The University of Melbourne, Victoria, Australia\\
    fan.jiang1@student.unimelb.edu.au, t.cohn@unimelb.edu.au
}

\begin{document}

\maketitle

\begin{abstract}
Syntax has been shown to benefit Coreference Resolution from incorporating long-range dependencies and structured information captured by syntax trees, either in traditional statistical machine learning based systems or recently proposed neural models. However, most leading systems use only dependency trees. We argue that constituent trees also encode important information, such as explicit span-boundary signals captured by nested multi-word phrases, extra linguistic labels and hierarchical structures useful for detecting anaphora. In this work, we propose a simple yet effective graph-based method to incorporate constituent syntactic structures. Moreover, we also explore to utilise higher-order neighbourhood information to encode rich structures in constituent trees. A novel message propagation mechanism is therefore proposed to enable information flow among elements in syntax trees. Experiments on the English and Chinese portions of OntoNotes 5.0 benchmark show that our proposed model either beats a strong baseline or achieves new state-of-the-art performance.\footnote{Code is available at https://github.com/Fantabulous-J/Coref-Constituent-Graph}
\end{abstract}

\section{Introduction}
As one of the most fundamental and important tasks in Natural Language Processing (NLP), Coreference Resolution aims to group all \textit{mentions} that refer to the same real-world entity. This is framed as a span-level classification problem over sequences of words, as illustrated in Figure~\ref{fig:coref_example}. 
Recently, neural models have achieved strong performance in coreference resolution~\citep{lee-etal-2018-higher,joshi-etal-2019-bert, joshi-etal-2020-spanbert,wu-etal-2020-corefqa} with the help of pretrained language models~\citep{peters-etal-2018-deep, devlin-etal-2019-bert}.

Syntax was widely used in early learning-based methods, through the use of features derived from syntax trees, which were shown to significantly improve statistical coreference systems~\citep{Ge98astatistical, bergsma-lin-2006-bootstrapping, kong-etal-2010-dependency, kong-zhou-2011-combining}. However, it is not clear whether these syntax-related features can improve modern neural coreference models. Meanwhile, many research in other tasks have got positive effects by introducing syntactic information into neural models. They normally apply graph neural networks~\citep{schlichtkrull2017modeling, velikovi2018graph} to automatically incorporate dependency trees to better capture the non-local relationships between words~\citep{marcheggiani-titov-2017-encoding, bastings-etal-2017-graph, schlichtkrull2017modeling}. A very recent work~\citep{jiang-cohn-2021-incorporating} shows that incorporating dependency syntax together with semantic role features using heterogeneous graph networks can significantly improve a strong neural coreference model. However, few attempts have been made for the application of constituent syntax, especially on neural coreference models.

\begin{figure}[!t]
\center
\begin{tabular}{p{8cm}}
\toprule
    \textit{DA:} It’s because of what \emph{\textcolor{blue}{both of you}} are doing to have things change. \underline{\textcolor{red}{I}} think that’s what’s… Go ahead Linda. \\
    \textit{LW:} Thanks goes to \underline{\textcolor{red}{you}} and to the media to help \emph{\textcolor{blue}{us}}. \\
    \textit{DA:} Absolutely. \\
    \textit{LW:} Obviously \emph{\textcolor{blue}{we}} couldn't seem loud enough to bring the attention, so \emph{\textcolor{blue}{our}} hat is off to all of you as well. \\
\bottomrule
\end{tabular}
\caption{An example of coreference resolution~\citep{pradhan-etal-2012-conll}. Coreferent mentions are with the same colour.}
\label{fig:coref_example}
\end{figure}

In this work, we aim to investigate the use of constituent syntax in neural coreference models, and we argue that compared to dependency syntax, incorporating constituent syntax is more natural for coreference resolution. In constituent trees, the information encoding boundaries of non-terminal phrases is explicitly presented. By contrast, such information is either implicitly encoded or not revealed in dependency trees. 
Besides, gold mentions usually map to a limited range of constituent types (See \S\ref{subsec:analysis}). We thus believe constituent tree structures capture signals for mention detection more effectively and explicitly, which we suspect is more helpful to the overall coreference resolution task. Besides, extra linguistic labels are also shown to reveal linguistic constraints for coreference resolution~\citep{ng-2010-supervised} (e.g., indefinite mentions are less likely to refer to any preceding mentions).

In order to effectively incorporate constituent syntax structures, we propose a graph-based neural coreference model. Previous works~\citep{trieu-etal-2019-coreference, ijcai2019-700}
introduced constituent trees as hard constraints to filter invalid mentions, which helps obtain better mention detectors and overall coreference resolvers. However, these methods do not preserve the full structure of original trees, as a result of their simplification of the tree as node traversal sequences with a collection of path features, or by ignoring the hierarchical structure entirely. In contrast, our method builds a graph consisting of terminal and non-terminal nodes and applies graph neural networks to encode such structures more flexibly.

Our model consists of three parts: document encoder, graph encoder and coreference layer. We use pretrained language model such as SpanBERT~\citep{joshi-etal-2020-spanbert} as document encoder. We propose a novel graph encoder by leveraging graph attention networks~\citep{velikovi2018graph}, which encodes bidirectional graphs separately. Moreover, in order to capture higher-order neighbourhood information and reduce the over-smoothing problem~\citep{Chen_Lin_Li_Li_Zhou_Sun_2020} of graph neural networks by stacking too many layers, we additionally add two-hop edges based on original constituent tree structures. New message propagation mechanisms over the underlying extended graph are therefore designed, where constituent nodes are updated iteratively using bidirectional graph attention networks, and explicit hierarchical syntax and span boundary information are propagated to enhance the contextualized token embeddings. We conduct experiments on the English and Chinese portions of OntoNotes 5.0 benchmark, and show that our proposed model significantly outperforms a strong baseline and achieves new state-of-the-art performance on the Chinese dataset.

\section{Baseline Model}\label{sec:appendix-baseline}
Our model is based on the \textsc{c2f-coref}\textit{+SpanBERT} model~\citep{joshi-etal-2019-bert}, which improves over ~\citet{lee-etal-2018-higher} with the document encoder replaced with SpanBERT model. It exhaustively enumerates all text spans up to a certain length limit as candidate mentions and aggressively prunes spans with low confidence. For each mention $i$, the model will learn a distribution over its possible antecedents $\mathcal{Y}(i)$:
\begin{equation}
    P(y)=\frac{e^{s(i,y)}}{\sum_{y'\in \mathcal{Y}(i)}e^{s(i,y')}}
\end{equation}
where the scoring function $s(i,j)$ measures how likely span $i$ and $j$ comprise valid mentions and corefer to one another:
\begin{align}
    s(i,j) &=s_{m}(i)+s_{m}(j)+s_{c}(i,j) \\
    s_{m}(i) &=\mathbf{FFNN}_{m}(\mathbf{g}_{\mathbf{i}}) \\
    s_{c}(i,j) &=\mathbf{FFNN}_{c}(\mathbf{g}_{\mathbf{i}}, \mathbf{g}_{\mathbf{j}}, \mathbf{g}_{\mathbf{i}}\odot\mathbf{g}_{\mathbf{j}}, \phi(i,j))
\end{align}
where $\mathbf{g}_{\mathbf{i}}$ and $\mathbf{g}_{\mathbf{j}}$ are span representations formed by the concatenation of contextualized embeddings of span endpoints, head vector using attention mechanism and span width embeddings. $\mathbf{FFNN}$ represents a two-layers feedforward neural network with \textbf{ReLU} activation function inside. $\mathbf{g}_{\mathbf{i}}\odot\mathbf{g}_{\mathbf{j}}$ is the element-wise similarity between span $i$ and $j$. $\phi(i,j)$ are meta features including bucket span distances, genre information and binary speaker features. $s_m$ means the predicted mention score which will be used to prune unlikely candidate mentions, while $s_c$ is the final pairwise coreference score. 
As suggested by~\citet{xu-choi-2020-revealing}, we further discard the higher-order based span refinement module, which uses antecedent distribution $P(y)$ as the attention mechanism to obtain the weighted average sum of antecedent embeddings, when replicating the baseline model.

\section{Proposed Model}\label{sec: cons-model}
\subsection{Document Encoder}\label{subsec: encoder}
In order to fit long documents, \citet{joshi-etal-2019-bert} chooses to split documents into independent segments. But the drawback is that it has limited modelling capacity as tokens can only attend to other tokens within the same segment, especially for tokens at the boundary of each segment~\citep{joshi-etal-2019-bert}. Instead, we choose to create overlapped segments and treat speakers (speaker's name of each utterance) as part of the input~\citep{wu-etal-2020-corefqa}. Specifically, we use a sliding-window approach to create $T$-sized segments with a stride of $\frac{T}{2}$ tokens and insert speakers into the beginning of their belonged utterances. Overlapped segments with attached speaker information are then encoded by SpanBERT~\citep{joshi-etal-2020-spanbert} to obtain contextualized representations, which can be denoted as $\mathbf{H}_w=(\mathbf{h}_1, \mathbf{h}_2, \dots, \mathbf{h}_n)$, where $\mathbf{h}_i\in\mathcal{R}^d$ and $n$ is the document length.

\begin{figure}[t]
    \centering
    \includegraphics[width=0.47\textwidth]{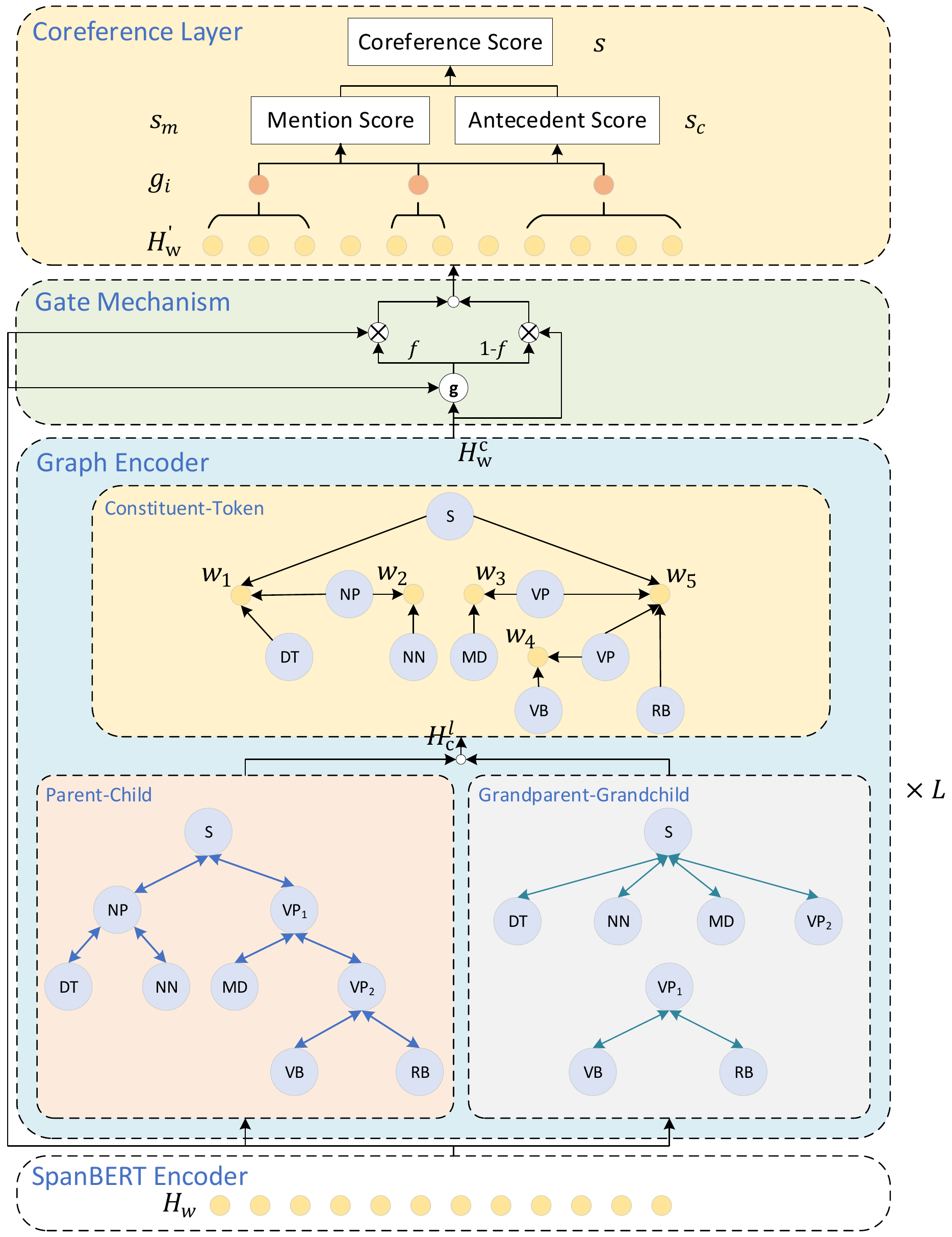}
    \caption{The architecture of our proposed model.}
    \label{architecture}
\end{figure}
\subsection{Graph Construction}
For each sentence in the document, we have an associated constituent tree which consists of words (terminals) and constituents (non-terminals). Therefore, we have two types of nodes in our graph: token nodes ($W=\{w_1,w_2,\dots,w_n\}$) and constituent nodes ($C=\{c_1,c_2,\dots,c_m\}$), where $n$ and $m$ are the number of token and constituent nodes, respectively. For a given constituent node $c_i$, we use START($c_i$) and END($c_i$) to denotes its start and end token indices.

\subsubsection{Node Initialization}\label{node-init}
Token node representations are initialized using the contextualized token representations $\mathbf{H}_w$ from the document encoder. Span boundary information has been shown to be extremely important to span-based tasks. Therefore, a novel span yield enhanced method is proposed to initialize each constituent node $c_i\in C$: 

\begin{gather}
    \mathbf{h}_{c_i}=[\mathbf{h}_{\text{START}(c_i)}; \mathbf{h}_{\text{END}(c_i)}; \mathbf{e}_{\text{type}}(c_i)]\label{cons_construct}
\end{gather}
where $\mathbf{h}_{\text{START}(c_i)}$ and $\mathbf{h}_{\text{END}(c_i)}$ are the embeddings of start and end tokens of constituent $c_i$. $\mathbf{e}_{\text{type}}(c_i)$ is the constituent type embeddings obtained from a randomly initialized look-up table, which will be optimized during the training process. Thus, we can obtain a set of initialized constituent node representations: $\mathbf{H}_c=\{\mathbf{h}_{c_1},\mathbf{h}_{c_2},\dots, \mathbf{h}_{c_m}\}$.

\subsubsection{Edge Construction}\label{subsec: edge-construct}
\paragraph{Constituent-Constituent} We design two categories of edges in our graph, namely \textit{parent-child} and \textit{grandparent-grandchild}, to capture longer-range dependencies. For each edge category, we further add reciprocal edges for each edge in the graph and label them with \textit{forward} and \textit{backward} types, respectively. Additionally, self-loop edges are added to each node in the graph. Thus, the edges are constructed based on following rules:

(1). A pair of \textbf{parent-child} and \textbf{child-parent} edges between node $c_i$ and $c_j$ are constructed if they are directly connected in the constituent tree. 

(2). A pair of \textbf{grandparent-grandchild} and \textbf{grandchild-grandparent} edges between node $c_i$ and $c_j$ are constructed if node $c_i$ can reach node $c_j$ using two hops, and vice versa.

\paragraph{Constituent-Token}
A token node $w_i$ is linked to $c_j$ if it is the left or rightmost token in the yield of $c_j$. Such edges are made unidirectional to make sure that information can only be propagated from constituent nodes to token nodes, which aims to enrich basic token representations with span boundary information and the hierarchical syntax structures.  

\subsection{Graph Encoder}
We use a Graph Attention Network~\citep{velikovi2018graph} to update the representation of constituent nodes and propagate syntactic information to basic token nodes. For a node $i$, the attention mechanism allows it to selectively incorporate information from its neighbour nodes:
\begin{align}
    \alpha_{ij} &= \text{softmax}(\sigma(\mathbf{a}^{T}[\mathbf{W} \mathbf{h}_{i}; \mathbf{W} \mathbf{h}_{j}]))\label{gat31}\\
    \mathbf{h}_{i}^{\prime}&=\|_{k=1}^{K} \text{ReLU}(\sum_{j} \alpha_{i j}^{k} \mathbf{W}^{k} \mathbf{h}_{j})\label{gat32}
\end{align}
where $K$ is the number of heads, $\mathbf{h}_i$ and $\mathbf{h}_j$ are embeddings of node $i$ and $j$, $\mathbf{a}^{T}$, $\mathbf{W}$ and $\mathbf{W}^{k}$ are trainable parameters. $\sigma$ is the LeakyReLU activation function~\citep{xu2015empirical}. $\|$ and $[;]$ means concatenation. Eqs.~\ref{gat31} and \ref{gat32} are designated as an operation $\mathbf{h}_i^\prime=\operatorname{GAT}(\mathbf{h}_i,\mathbf{h}_j|j\in\mathcal{N}_i)$, where $\mathcal{N}_i$ is the set of target node $i$'s neighbours, $\mathbf{h}_i$ and $\mathbf{h}_j$ are the embeddings of target and neighbour node. $\mathbf{h}_i^\prime$ is the updated embedding of target node. 

\paragraph{Bidirectional GAT Layer}
We design a bidirectional GAT layer to model the constituent-constituent edges. Specifically, for a given constituent node $c_i$, we obtain its neighbour nodes with edge type $t$ in \textit{forward} (outgoing) and \textit{backward} (incoming) directions: $\mathcal{N}_{c_i}^{tf}$ and $\mathcal{N}_{c_i}^{tb}$, respectively. Then we use two separate GAT encoders to derive the updated representation of node $c_i$ in different directions:
\begin{align}
    \mathbf{h}_{c_i}^{tf} &= \operatorname{GAT}(\mathbf{h}_{c_i},{\mathbf{h}_{c_j} | c_j \in \mathcal{N}_{c_i}^{tf}})\label{multi-gat1}\\
    \mathbf{h}_{c_i}^{tb} &= \operatorname{GAT}(\mathbf{h}_{c_i},{\mathbf{h}_{c_j} | c_j \in \mathcal{N}_{c_i}^{tb}})
\end{align}
Then the updated representation of constituent node $c_i$ is obtained by the summation of the representations of two directions: $\mathbf{h}_{c_i}^{t} = \mathbf{h}_{c_i}^{tf} + \mathbf{h}_{c_i}^{tb}$.

\paragraph{Multi-type Integration Layer}
In order to aggregate updated node representations using different types of edges, we use the self-attentive mechanism~\cite{lee-etal-2017-end}:
\begin{align}
    \alpha_{c_i, t}&= \operatorname{softmax}(\textbf{FFNN}(\mathbf{h}_{c_i}^{t})) \\
    \mathbf{h}_{c_i}^\prime&=\sum_{t=1}^{T}\alpha_{c_i, t} \mathbf{h}_{c_i}^{t}\label{multi-gat2}
\end{align}
where $T$ is the number of edge types and \textbf{FFNN} is a two-layer feedforward neural network with \textbf{ReLU} function inside. An operation is designated to summarise Eqs.~\ref{multi-gat1} to \ref{multi-gat2}: \mbox{$\mathbf{h}_{c_i}^\prime = \operatorname{Multi-BiGAT}(\mathbf{h}_{c_i}, \mathbf{h}_{c_j}|c_j\in \mathcal{N}_{c_i})$}.

\subsection{Message Propagation}
The message propagation mechanism is defined to enable information flow from constituent nodes to basic token nodes. First, we update the constituent node representation using our defined bidirectional GAT layer with multi-type edges:
\begin{gather}
    \mathbf{h}_{c_i}^{l+1} = \operatorname{Multi-BiGAT}(\mathbf{h}_{c_i}^{l}, \mathbf{h}_{c_j}^{l}|c_j\in \mathcal{N}_{c_i})
\end{gather}
where $\mathbf{h}_{c_i}^{l}$ is the constituent node representation from previous layer $l$ and $\mathbf{h}_{c_i}^0$ is the initial constituent node embedding.

Then the updated constituent node representations propagate information to update the token node embeddings through constituent-token edges:
\begin{gather}
    \mathbf{h}_i^{l+1}=\operatorname{GAT}(\mathbf{h}_i^{l},\mathbf{h}_{c_j}^{l+1}|c_j\in\mathcal{N}_i)
\end{gather}
where $\mathbf{h}_i^{l}$ is the token representation from layer $l$ and $\mathbf{h}_i^0$ is the encoding from document encoder. 

The updated token representation is then used to reconstruct the updated constituent node embeddings using Eq.~\ref{cons_construct}, which will be employed in the next graph encoder layer. After $L$ iterations, we could obtain the final constituent syntax enhanced token representations, which are denoted as $\mathbf{H}_w^c$.

Finally, we use a gating mechanism to infuse the syntax-enhanced token representation dynamically:
\begin{align}
    \mathbf{f} &= \sigma(\mathbf{W}_g\cdot [\mathbf{H}_w; \mathbf{H}_w^c]+\mathbf{b}_g)\\
    \mathbf{H}^\prime_w &= \mathbf{f} \odot \mathbf{H}_w + (\mathbf{1}-\mathbf{f}) \odot \mathbf{H}_w^c
\end{align}
where $\mathbf{W}_g$ and $\mathbf{b}_g$ are trainable parameters, $\odot$ and $\sigma$ are element-wise multiplication and sigmoid function respectively.

The enhanced token representations $\mathbf{H}_w^\prime$ will be used to form span embeddings and compute coreference scores~\citep{lee-etal-2018-higher} (See \S\ref{sec:appendix-baseline}).

\begin{table*}[t]
\footnotesize
\centering
\begin{tabular}{l|l@{\hspace{0.45cm}}c@{\hspace{0.25cm}}c@{\hspace{0.25cm}}c@{\hspace{0.45cm}}c@{\hspace{0.25cm}}c@{\hspace{0.25cm}}c@{\hspace{0.45cm}}c@{\hspace{0.25cm}}c@{\hspace{0.25cm}}c@{\hspace{0.45cm}}c }
    \toprule
    \multicolumn{2}{c}{} & \multicolumn{3}{c}{\textbf{MUC}} & \multicolumn{3}{c}{\textbf{$\text{B}^3$}}& \multicolumn{3}{c}{\textbf{$\text{CEAF}_{\phi_4}$}} & \\
    \multicolumn{2}{c}{} & P & R & F1 & P & R & F1 & P & R & F1 & Avg. F1 \\
    \toprule
    \multirow[c|]{12}{*}{English} &
    \textsc{e2e-coref}~\cite{lee-etal-2017-end} & 78.4 & 73.4 & 75.8 & 68.6 & 61.8 & 65.0 & 62.7 & 59.0 & 60.8 & 67.2 \\
    & \textsc{c2f-coref}~\cite{lee-etal-2018-higher} & 81.4 & 79.5 & 80.4 & 72.2 & 69.5 & 70.8 & 68.2 & 67.1 & 67.6 & 73.0 \\
    \cmidrule{2-12}
    & SpanBERT-base~\citep{joshi-etal-2020-spanbert} & 84.3 & 83.1 & 83.7 & 76.2 & 75.3 & 75.8 & 74.6 & 71.2 & 72.9 & 77.4\\
    & Our baseline + SpanBERT-base$^{\ast\dagger}$ & 83.9 & 84.2 & 84.0 & 76.2 & 76.9 & 76.6 & 74.3 & 73.1 & 73.7 & 78.1 ($\pm$0.1)\\
    & \citet{jiang-cohn-2021-incorporating} + SpanBERT-base$^{\dagger\ddag\S}$ & 85.3 & 85.0 & 85.2 & 77.9 & 77.7 & 77.8 & 75.6 & 74.1 & 74.8 & 79.3 ($\pm$0.2)\\
    & \textbf{Our model} + SpanBERT-base$^{\dagger\S}$ & \textbf{85.6} & 85.8 & 85.7 & 78.2 & \textbf{79.0} & \textbf{78.6} & \textbf{76.3} & 74.8 & 75.5 & \textbf{80.0} ($\pm$0.2) \\
    & CorefQA~\citep{wu-etal-2020-corefqa} + SpanBERT-base$^{\P}$ & 85.2 & \textbf{87.4} & \textbf{86.3} & \textbf{78.7} & 76.5 & 77.6 & 76.0 & \textbf{75.6} & \textbf{75.8} & 79.9 \\
    \cmidrule{2-12}
    & SpanBERT-large~\citep{joshi-etal-2020-spanbert} & 85.8 & 84.8 & 85.3 & 78.3 & 77.9 & 78.1 & 76.4  & 74.2 & 75.3 & 79.6 \\
    & Our baseline + SpanBERT-large$^{\ast\dagger}$ & 86.0 & 86.0 & 86.0 & 79.6 & 79.6 & 79.6 & 77.2 & 75.8 & 76.5 & 80.7 ($\pm$0.1) \\
    & \citet{jiang-cohn-2021-incorporating} + SpanBERT-large$^{\dagger\ddag\S}$ & 87.2 & 86.7 & 87.0 & 81.1 & 80.5 & 80.8 & 78.6 & 77.0 & 77.8 & 81.8 ($\pm$0.2) \\
    & \textbf{Our model} + SpanBERT-large$^{\dagger\S}$ & 87.3 & 87.1 & 87.2 & 81.1 & 80.9 & 81.0 & 78.8 & 77.2 & 78.0 & 82.1 ($\pm$0.2)\\
    & CorefQA~\cite{wu-etal-2020-corefqa} + SpanBERT-large$^{\P}$ & \textbf{88.6} & \textbf{87.4} & \textbf{88.0} & \textbf{82.4} & \textbf{82.0} & \textbf{82.2} & \textbf{79.9}  & \textbf{78.3} & \textbf{79.1} & \textbf{83.1} \\
    \toprule
    \multirow{6}{*}{Chinese} & 
    \citet{clark-manning-2016-improving} & 73.9 & 65.4 & 69.4 & 67.5 & 56.4 & 61.5 & 62.8 & 57.6 & 60.1 & 63.7 \\
    & \citet{ijcai2019-700}$^{\S}$ & 77.0 & 64.6 & 70.2 & 70.6 & 54.7 & 61.6 & 64.9 & 55.4 & 59.8 & 63.9 \\
    & Our baseline + BERT-wwm-base$^{\ast\dagger}$ & 76.7 & 70.9 & 73.7 & 68.3 & 62.4 & 65.2 & 67.4 & 60.8 & 63.9 & 67.6 ($\pm$0.3) \\
    & \textbf{Our model} + BERT-wwm-base$^{\dagger\S}$ & \textbf{84.1} & \textbf{78.6} & \textbf{81.3} & \textbf{77.4} & \textbf{71.5} & \textbf{74.4} & \textbf{76.5} & \textbf{70.0} & \textbf{73.1} & \textbf{76.3} ($\pm$0.2) \\
    \cmidrule{2-12}
    & Our baseline + RoBERTa-wwm-ext-large$^{\ast\dagger}$ & 79.9 & 72.2 & 75.8 & 71.6 & 64.3 & 67.7 & 70.8 & 62.8 & 66.5 & 70.0 ($\pm$0.3) \\
    & \textbf{Our model} + RoBERTa-wwm-ext-large$^{\dagger\S}$ & \textbf{85.8} & \textbf{80.9} & \textbf{83.3} & \textbf{79.8} & \textbf{74.5} & \textbf{77.0} & \textbf{78.7} & \textbf{72.9} & \textbf{75.7} & \textbf{78.7} ($\pm$0.2) \\
    \bottomrule
    \end{tabular}
    \caption{The results on the test set of the OntoNotes English and Chinese shared task compared with previous systems when using gold constituent trees. $\ast$ indicates our replicated baseline. $\S$ indicates methods using gold features. $\P$ indicates methods using substantial training resources and extra datasets for pretraining. $\dagger$ means averaged performance over 5 runs. $\ddag$ means results obtained by running the publicly released code of~\citet{jiang-cohn-2021-incorporating} with the document encoding method in Section 3.1.}
\label{tab:results}
\end{table*}

\section{Experiments}

\subsection{Dataset}
Our model is evaluated on the English and Chinese portions of OntoNotes 5.0 dataset~\citep{pradhan-etal-2012-conll}. The English corpus consists of 2802, 343 and 348 documents in the training, development and test splits, respectively, while the Chinese corpus contains 1810, 252 and 218 documents for train/dev/test splits.
The model is evaluated using three coreference metrics: MUC, B$^3$ and CEAF$\phi_4$ and the average F1 score (Avg. F1) of the three are reported. We use the latest version of the official evaluation scripts (version 8.01),\footnote{http://conll.cemantix.org/2012/software.html} which implements the original definitions of the metrics.

\subsection{Experimental Settings}
We reimplement the \textsc{c2f-coref}\textit{+SpanBERT}\footnote{https://github.com/mandarjoshi90/coref} baseline using PyTorch. For English model, we use SpanBERT-base and large model to encode documents;\footnote{https://github.com/facebookresearch/SpanBERT} while for Chinese, we use BERT-wwm-base and RoBERTa-wwm-ext-large\footnote{https://github.com/ymcui/Chinese-BERT-wwm} as the document encoders. Graph attention networks and the message propagation module are implemented based on Deep Graph Library~\citep{DBLP:journals/corr/abs-1909-01315}. Gold constituent trees annotated on the datasets are used in this experiment for consistent comparison with previous work.

Most hyperparameters are adopted from \citet{joshi-etal-2019-bert} and newly introduced hyperparameters are determined through grid search. The learning rates of finetuning base and large model are $2\times 10^{-5}$ and $1\times 10^{-5}$. The learning rates of task-specific parameters are $3\times 10^{-4}$ and $5\times 10^{-4}$ for English, and $5\times 10^{-4}$ for Chinese when using base and large model, respectively. Both BERT and task parameters are trained using Adam optimizer~\citep{DBLP:journals/corr/KingmaB14}, with a warmup learning scheduler for the first 10\% of training steps and linear decay scheduler decreasing to 0, respectively. The number of heads is set to 4 and 8 for base and large models. The size of constituent type embeddings is 300. We set the number of graph attention layers as 2. For all experiments, we choose the best model according to Avg. F1 on dev set, which is then evaluated on the test set.

\subsection{Baselines and State-of-the-Art}
We compare our proposed model with a variety of previous competitive models: \citet{clark-manning-2016-improving} is a neural network based model which incorporates entity-level information. \textsc{e2e-coref}~\citep{lee-etal-2017-end} is the first end-to-end neural model for coreference resolution which jointly detects and groups entity mention spans. \citet{ijcai2019-700} improves the \textsc{e2e-coref} model by treating constituent trees as constraints to filter invalid candidate mentions and encoding the node traversal sequences of parse trees to enhance document representations. \textsc{c2f-coref}~\citep{lee-etal-2018-higher} extends the \textsc{e2e-coref} model by introducing a \emph{coarse-to-fine} candidate mention pruning strategy and a higher-order span refinement mechanism. \citet{joshi-etal-2020-spanbert} improves over \textsc{c2f-coref} with the document encoder replaced by SpanBERT. CorefQA~\citep{wu-etal-2020-corefqa} employs the machine reading comprehension framework to recast the coreference resolution problem as a query-based span-prediction task, which achieves current state-of-the-art performance. \citet{jiang-cohn-2021-incorporating} enhances neural coreference resolution by incorporating dependency syntax and semantic role labels using heterogeneous graph attention networks.

\subsection{Overall Results}
Table~\ref{tab:results} shows the results of our model compared with a range of high-performing neural coreference resolution models on English and Chinese. For English, we observe that our replicated baseline surpasses the SpanBERT baseline~\citep{joshi-etal-2020-spanbert} by 0.7\% and 1.1\%, demonstrating the effectiveness of the sliding-window based document encoding approach and modified representations of speaker identities. Our model further improves the replicated baseline significantly with improvements of 1.9\% and 1.4\%, respectively, a result which is also comparable to the state-of-the-art performance of CorefQA~\citep{wu-etal-2020-corefqa}.\footnote{We do not use their model as a baseline mainly due to hardware limitations, as it requires 128G GPU memory for training. It can also be easily incorporated with our method by adding the proposed graph encoder on top of their document encoder with minor modification, which we expect would lead to further improvements.} Improvements can also be observed over \citet{jiang-cohn-2021-incorporating} (0.7\% with $p<$ 0.002 and 0.3\% with $p<$ 0.06).\footnote{Pitman’s permutation statistical test~\citep{dror-etal-2018-hitchhikers}.} For Chinese, our replicated baseline has already achieved state-of-the-art performance. With the help of constituent syntax, our model again beats the baseline model with significant improvements of 8.7\%. This indicates that constituent syntax is far more useful to Chinese than English, and we suspect that word-level segmentation encoded in constituent trees brings extra benefits in Chinese. 

\subsection{Analysis}\label{subsec:analysis}
\noindent\textbf{Effects of Constituency Quality}
To evaluate how the quality of constituent trees affects the performance, we test two off-the-shelf parsers~\citep{zhang-etal-2020-fast} (achieving 95.26\% and 91.40\% F1 score on PTB and CTB7) to obtain predicted trees. When using predicted trees with our base model, we get Avg. F1 of 78.7\% (+0.6\% with $p<$ 0.05) and 73.0\% (+5.4\%) on both languages, consistently outperforming the baseline. Similarly, the effects in large models are also noticeable, resulting in Avg. F1 of 81.0\% (+0.3\% with $p<$ 0.05) and 75.0\% (+5.0\%) respectively. However, the performance is still worse than using gold trees, indicating the necessity of high-quality constituency parsers. 

\noindent\textbf{Ablation Study}
We modify several components of our model to validate their effects. Results are reported in Table~\ref{tab:ablation} for the following ablations: 1) Using vanilla constituent trees by only keeping parent-child edges; 2) Removing the gating mechanism and directly use representations from the graph encoder; 3) Changing the way of representing constituent node to initialize only with type embeddings; 4) Using the \textit{independent} setting~\citep{joshi-etal-2019-bert} for document encoding; 5) Using the similar way as in this paper to incorporate dependency trees; 6) Using the same method to encode constituent and dependency syntax as in this paper alongside the method in~\citet{jiang-cohn-2021-incorporating} for attention fusion of the dependency-syntax and constituent-syntax representations.

From Table~\ref{tab:ablation} we observe that: 1) The bidirectional graph and higher-order edges show positive impacts in capturing long-range dependencies; 2) Removing the gate mechanism leads to significant performance degradation, especially in English. We believe that the gate mechanism plays an important role in dynamically choosing useful information from original sequential representations and graph-enhanced representations, and keeping information such as position embeddings from being lost after the graph attention network; 3) Although only using type embeddings to initialize constituent node representations also yields competitive performance, our span yield enhanced initialization method can capture span-boundary information more effectively; 4) Splitting documents into independent segments is less beneficial, especially for tokens at the boundary of their segment; 5) Incorporating dependency syntax achieves inferior performance, showing that explicit span-boundary information encoded in constituent trees is more beneficial; 6) Combining both types of syntax is better than using dependency syntax solely but is inferior to only using constituent syntax.

\begin{table}[t]
    \footnotesize
    \centering
    \begin{tabular}{l@{\hspace{0.35cm}}c@{\hspace{0.2cm}}c@{\hspace{0.35cm}}c@{\hspace{0.2cm}}c}
        \toprule
        & \multicolumn{2}{c}{English} & \multicolumn{2}{c}{Chinese} \\
        Variants & Avg. F1 & $\Delta F$1 & Avg. F1 & $\Delta F$1 \\
        \midrule
        - & 80.0 & - & 76.3 & - \\
        Vanilla Tree & 79.7 & -0.3 & 75.9 & -0.4 \\
        No Gate & 77.3 & -2.7 & 75.1 & -1.2 \\
        Only Type Embedding & 79.5 & -0.5 & 75.5 & -0.8 \\
        No Sliding Window & 79.6 & -0.4 & 76.0 & -0.3 \\
        Dependency Syntax & 79.2 & -0.8 & 70.8 & -5.5 \\
        Constituent \& Dependency & 79.7 & -0.3 & 75.3 & -1.0 \\
        \bottomrule
    \end{tabular}
    \caption{Results when ablating different modules compared to our base model on English and Chinese datasets.}
    \label{tab:ablation}
\end{table}
\begin{table}[t]
    \footnotesize
    \centering
    \begin{tabular}{l@{\hspace{0.25cm}}l@{\hspace{0.25cm}}c@{\hspace{0.25cm}}c@{\hspace{0.25cm}}c@{\hspace{0.25cm}}c@{\hspace{0.25cm}}c@{\hspace{0.25cm}}c}
        \toprule
        \multirow{2}{*}{Dataset} & \multirow{2}{*}{Model} & \multicolumn{5}{c}{Mention Length} & \multirow{2}{*}{Overall} \\
        & & 1-2 & 3-4 & 5-7 & 8-10 & 11+ & \\
        \midrule
        \multirow{2}{*}{English} & baseline & 90.3 & 82.8 & 78.3 & 75.1 & 65.9 & 87.0 \\
        & our method & \textbf{90.5} & \textbf{84.3} & \textbf{80.1} & \textbf{79.2} & \textbf{74.0} & \textbf{88.1} \\
        \midrule
        \multirow{2}{*}{Chinese} & baseline & 85.1 & 79.0 & 71.0 & 69.1 & 66.4 & 80.1 \\
        & our method & \textbf{88.5} & \textbf{85.0} & \textbf{78.3} & \textbf{80.0} & \textbf{73.6} & \textbf{85.1} \\
        \bottomrule
    \end{tabular}
    \caption{The F1 score based on mention length on English and Chinese development sets when using base model.}
    \label{tab:mention-length}
\end{table}

\noindent\textbf{Mentions With Different Lengths}
Table~\ref{tab:mention-length} shows the performance comparison in terms of different mention lengths on both datasets. As shown in the table, we can observe that our proposed model consistently outperforms the baseline model for both two languages. This indicates that the improved overall performance in the coreference resolution task has benefited largely from better mention detectors, which is consistent with our hypothesis. The performance gain is more significant for mentions with longer length on both languages, demonstrating that leveraging constituent syntax is highly effective for modelling long-range dependencies and becomes more crucial when entity length becomes longer. 

\noindent\textbf{Document Length}
In Table~\ref{tab:document_length}, we show that the performance of our model against the baseline on the English development set as a function of document lengths. As expected, our model consistently outperforms the baseline model on all document sizes, especially for documents with length larger than 1153 tokens. This demonstrates that the incorporated constituent syntax and our modelling choices are beneficial for capturing longer-range dependencies. Besides, the improvements on short documents ($<$128 tokens) are also significant. We find that most anaphoric mentions have very short distances between their nearest antecedents. The Binding Theory~\citep{chomsky_language_1988} argues that constituent syntax is more effective in keeping anaphoric mentions locally bounded by short-distance antecedents. Thus, it is possible that our model implicitly learns this principle, which results in better performance. Nevertheless, our model shows similar pattern as the baseline model, performing distinctly worse as document length increases. This indicates that the sentence-level syntax used in this work are not sufficient enough to tackle the deficiency of modelling long-range dependency. One possible solution is to incorporate document-level features such as hierarchical discourse structures.

\begin{table}[t]
    \footnotesize
    \centering
    \begin{tabular}{r@{\hspace{0.3cm}}c@{\hspace{0.3cm}}c@{\hspace{0.3cm}}c@{\hspace{0.3cm}}c}
        \toprule
         Doc length & \#Docs & Baseline & Ours & $+\Delta F$1  \\
         \midrule
         0 -- 128 & 57 & 82.8 & 86.1 & +3.3 \\
         129 -- 256 & 73 & 81.3 & 83.2 & +1.9 \\
         257 -- 512 & 78 & 82.0 & 83.7 & +1.7 \\
         513 -- 768 & 71 & 78.3 & 79.0 & +0.7 \\
         769 -- 1152 & 52 & 77.5 & 78.9 & +1.4 \\
         1153+ & 12 & 68.0 & 70.7 & +2.7 \\
         \midrule
         All & 343 & 78.2 & 79.7 & +1.5 \\
        \bottomrule
    \end{tabular}
    \caption{The Avg. F1 on the English dev set when using base model, broken down by document length.}
    \label{tab:document_length}
\end{table}

\begin{table}[t]
    \footnotesize
    \centering
    \begin{tabular}{ll@{\hspace{0.5cm}}c@{\hspace{0.3cm}}c}
        \toprule
        Dataset & Methods & Avg. F1 & $\Delta F$1 \\
        \midrule
        \multirow{3}{*}{English} & Baseline & 78.1 & - \\ 
        & Our Method & 80.0 & +1.9 \\
        & Baseline + Mention Filter & 77.3 & -0.8 \\
        \midrule
        \multirow{3}{*}{Chinese} & Baseline & 67.6 & - \\
        & Our Method & 76.3 & +8.7 \\
        & Baseline + Mention Filter & 71.8 & +4.2 \\
        \bottomrule
    \end{tabular}
    \caption{Results when utilising syntactic parse trees as mention filter compared to the baseline and our base model.}
    \label{tab:mention-filter}
\end{table}

\noindent\textbf{Constituent Tree as Mention Filter}
An alternative use of syntax is through constraining mention types. We use the constituent parse tree as hard constraints on top of the baseline to filter out invalid candidate mentions, assuming that only candidate mentions that have matched phrases in the parse tree are valid. We observe that about 99\% of gold mentions correspond to a small set of syntactic phrases and POS types.\footnote{en: the set of phrases tagged with NP, NML, PRP, PRP\$, WP, WDT, WRB, NNP, VB, VBD, VBN, VBG, VBZ, VBP~\citep{wu2020understanding} includes 99.63\% gold mentions. zh: the set of VV, NT, PN, DFL, NR, NP, QP, NN covers 99.79\% gold mentions.} We thus use these two phrase sets as filters to prune unlikely candidate mentions. Table~\ref{tab:mention-filter} shows the corresponding results. We can find that the syntactic constraint harms the performance slightly on the English baseline (-0.8\%) but improves the Chinese baseline by 4.2\%. However, in both cases this constrained baseline is substantially worse than using the syntax tree as part of our neural model, as proposed in this paper (with scores of 2.7\% and 4.5\% lower for English and Chinese, respectively).

\begin{table*}[t]
    \footnotesize
    \centering
    \begin{tabular}{l|c|c@{\hspace{0.5cm}}c|c@{\hspace{0.3cm}}c|c|c@{\hspace{0.5cm}}c|c@{\hspace{0.3cm}}c}
        \toprule
         & \multicolumn{5}{c|}{OntoNotes 5.0 English} & \multicolumn{5}{c}{OntoNotes 5.0 Chinese} \\
         \midrule
         & & \multicolumn{2}{c|}{Baseline} & \multicolumn{2}{c|}{Ours} & & \multicolumn{2}{c|}{Baseline} & \multicolumn{2}{c}{Ours}  \\
         Class & Size \% & RA & MD & RA & MD & Size \% & RA & MD & RA & MD \\
         \midrule
         PN-e & 15.52 & 96.4 & 94.0 & \textbf{96.5} & \textbf{94.3} &
         15.05 & \textbf{97.5} & 93.5 & 97.3 & 95.9 \\
         PN-p & 6.06 & 90.0 & 86.3 & \textbf{92.6} & \textbf{89.1} &
         2.81 & 76.0 & 84.7 & \textbf{84.7} & \textbf{88.2} \\
         PN-n & 6.63 & 85.7 & 89.1 & \textbf{86.4} & \textbf{89.5} &
         0.47 & 35.9 & 59.1 & \textbf{44.2} & \textbf{65.2} \\
         PN-na & 6.74 & 95.5 & 83.0 & \textbf{95.9} & \textbf{85.8} &
         6.05 & 85.1 & 78.4 & \textbf{85.7} & \textbf{85.4} \\
         \midrule
         CN-e & 6.17 & 96.7 & 90.8 & \textbf{96.9} & \textbf{91.9} & 
         16.01 & 91.5 & 80.1 & \textbf{93.5} & \textbf{85.8} \\
         CN-p & 8.60 & 83.8 & 80.5 & \textbf{85.9} & \textbf{82.3} &
         11.99 & 70.0 & 65.1 & \textbf{76.2} & \textbf{71.7} \\
         CN-n & 3.39 & 72.4 & 68.6 & \textbf{73.6} & \textbf{70.7} & 
         4.55 & 62.8 & 68.7 & \textbf{66.2} & \textbf{73.2} \\
         CN-na & 15.47 & 91.8 & 69.7 & \textbf{92.6} & \textbf{72.2} & 
         26.88 & \textbf{67.1} & 65.2 & 65.9 & \textbf{74.3} \\
         \midrule
         PR-1/2 & 11.64 & 93.8 & \textbf{95.1} & \textbf{93.9} & 94.9 &
         8.12 & 85.7 & \textbf{97.5} & \textbf{88.0} & 95.7 \\
         PR-G3 & 5.99 & 96.5 & 99.6 & \textbf{97.0} & \textbf{99.7} & 
         4.91 & 87.9 & \textbf{99.4} & \textbf{89.0} & 99.3 \\
         PR-UG3 & 10.14 & 86.8 & \textbf{95.2} & \textbf{88.8} & 94.8 & 
         1.02 & 72.5 & 91.0 & \textbf{73.9} & \textbf{95.8} \\
         PR-oa & 1.45 & 63.0 & \textbf{68.2} & \textbf{67.8} & 63.9 & 
         0.83 & \textbf{70.7} & \textbf{63.6} & 70.2 & 48.3 \\
         PR-na & 2.20 & 54.7 & \textbf{85.3} & \textbf{57.4} & 85.1 & 
         1.33 & 62.4 & \textbf{87.8} & \textbf{70.4} & 84.6 \\
        \bottomrule
    \end{tabular}
    \caption{The results of resolution classes in the development set of OntoNotes English and Chinese dataset.}
    \label{tab:resolution_class}
\end{table*}

\subsection{Resolution Classes}\label{subsec:resolution_classes}
To further understand the behaviour of our proposed model, we follow~\citet{stoyanov-etal-2009-conundrums} and \citet{lu-ng-2020-conundrums} to classify gold mentions into different resolution classes and compare it with the baseline on each of them.

\noindent\textbf{Proper Names}
Gold mentions associated with named entity types belong to this class, and four sub-classes are defined accordingly. 1) \textit{exact string match} (\emph{e}): at least one preceding mention in a proper name's gold cluster exactly has the same string; 2) \textit{partial string match} (\emph{p}): at least one preceding mention in a proper name's gold cluster shares some words; 3) \textit{no string match} (\emph{n}): no preceding mention in a proper name's gold cluster shares some words; 4) \textit{non-anaphoric} (\emph{na}): a proper name does not refer to any preceding mention. 

\noindent\textbf{Common NPs}
Gold mentions without named entity types belong to this class, with four sub-classes as in \textit{proper names}. 

\noindent\textbf{Pronouns}
Five pronoun sub-classes are defined. 1) \emph{1/2}: 1st and 2nd person pronouns (e.g., you); 2) \emph{G3}: gendered 3rd person pronouns (e.g., she); 3) \emph{U3}: ungendered 3rd person pronouns (e.g., they); 4) \emph{oa}: any anaphoric pronouns not in 1), 2), and 3) (e.g., demonstrative pronouns); 5) \emph{na}: non-anaphoric pronouns (e.g., pleonastic pronouns).

\noindent\textbf{Results}
For performance measurements, we follow \citet{lu-ng-2020-conundrums} to use mention detection recall (MD) and resolution accuracy (RA). For MD, we count the percentage of gold mentions that are correctly detected in each resolution class; while for RA, we compute the percentage of correctly detected mentions that are correctly resolved.

Table~\ref{tab:resolution_class} shows the performance of the baseline and our proposed model on each resolution class. Firstly, we can see that both models perform the best on proper names, followed by common nouns and pronouns. Secondly, by analysing the fine-grained classes, the \textit{exact match} class in proper names and common nouns are easier than the \textit{partial match} one, which is easier than the \textit{no string match} class. For pronouns, the 3rd person gendered pronoun is the easiest one, followed by the 1st/2nd person noun, while both models find it difficult to resolve other pronouns such as reflective pronouns.
Thirdly, we find that our model gains significant improvements on non-anaphoric mentions, showing its superiority in dealing with the difficulty of anaphoricity determination, with improvements up to 2.7\% and 8.0\% RA in English and Chinese, respectively. Moreover, considerable improvements on 3rd ungendered pronouns (2.0\% and 1.4\%) are also observed. Constituent syntax is also especially helpful in detecting partial and no string match classes for proper names (8.7\% and 8.3\%) and common NPs (6.2\% and 3.4\%) in Chinese. These demonstrate that the harder a resolution class is, the more significant our model's improvement is. Besides, this also shows that the incorporated constituent syntax help resolve traditionally difficult anaphors. Overall, by maintaining comparable performance in other easier classes simultaneously, our model has achieved significantly better final results on these two languages compared with the baseline.

\section{Related Work}

\paragraph{Syntax for Coreference Resolution}
Syntactic features derived from syntactic parse trees were dominant in early research for coreference resolution. \citet{Ge98astatistical} proposes Hobbs distances to encode the rank of candidate antecedents of a given pronoun based on Hobbs's syntax parse tree based pronoun resolution algorithm~\citep{hobbs-resolving-1987}. \citet{bergsma-lin-2006-bootstrapping} implements path-related features based on syntactic parse trees, where the sequence of words and dependency labels in the path between a given pronoun and its candidate antecedent is utilised. Statistical information collected from such paths is used to measuring the likelihood of being coreferent for the pronoun and antecedent. Syntactic information has also been applied in the anaphoricity determination task by using tree-kernel-based methods: \citet{kong-etal-2010-dependency} and \citet{kong-zhou-2011-combining} design various kinds of path-related features such as root path between the root node and current mention. By contrast, few attempts have been made to evaluate the utility of syntax for neural coreference models. \citet{trieu-etal-2019-coreference} and \citet{ijcai2019-700} treated constituent trees as signals to filter invalid candidate mentions for coreference resolution. Nevertheless, their methods either ignore the hierarchical structures of constituent trees or fail to well preserve tree structures by encoding constituent trees using node traversal sequences. To fill this gap, we propose a novel graph-based method to fully model constituent tree structures more flexibly.

\paragraph{Enhancing Neural Models with Syntax}
External syntax has long been used for enhancing neural models. Mainstream methods typically use graph neural networks to capture the structural information encoded in dependency trees. \citet{marcheggiani-titov-2017-encoding} and \citet{bastings-etal-2017-graph} applied Graph Convolutional Networks (GCNs)~\citep{schlichtkrull2017modeling} to incorporate dependency trees to capture the non-local relationships between words. \citet{wang-etal-2020-relational} employed reshaped dependency trees using relational graph attention networks to effectively capture long-range dependencies while ignoring noisy relations. 

Compared to dependency syntax, the utility of constituent syntax in neural models is less well studied. Some early work utilised recursive neural networks to incorporate constituent trees by recursively updating the representation of constituent phrases~\citep{socher-etal-2013-recursive, tai-etal-2015-improved}. Nevertheless, such a method is less efficient than applying graph neural networks since the recursive way of encoding means that later steps should depend on earlier ones.
The most similar method to our own is \citet{marcheggiani-titov-2020-graph}, who developed a neural model of semantic role labelling based on GCN encoding of a constituency tree with a message propagation mechanism. Nevertheless, our method differs in extending plain parse trees with higher-order edges and bidirectional graphs to capture longer-range neighbourhood information. We also proposes a novel span yield enhanced method to represent constituent nodes instead of initializing them with zero vectors, which is better suited to conference resolution and similar to the way of representing mention spans. Our work also differs in terms of the task: we consider coreference resolution rather than SRL, a document-level task requiring modelling inter-sentence phenomena.

\section{Conclusion}
In this paper, we successfully incorporated constituent trees with added higher-order edges and bidirectional graphs, which are encoded via our designed bidirectional graph attention networks and message propagation mechanism. Emperical results on a English and Chinese benchmark confirm the superiority of our proposed method, significantly beating a strong baseline and achieving state-of-the-art performance.

\section*{Acknowledgments}
We thank the anonymous reviewers for their helpful feedback. This research was undertaken using the LIEF HPC-GPGPU Facility hosted at the University of Melbourne. This Facility was established with the assistance of LIEF Grant LE170100200.

\bibliography{aaai22}

\begin{thebibliography}{37}
\providecommand{\natexlab}[1]{#1}

\bibitem[{Bastings et~al.(2017)Bastings, Titov, Aziz, Marcheggiani, and
  Sima{'}an}]{bastings-etal-2017-graph}
Bastings, J.; Titov, I.; Aziz, W.; Marcheggiani, D.; and Sima{'}an, K. 2017.
\newblock Graph Convolutional Encoders for Syntax-aware Neural Machine
  Translation.
\newblock In \emph{Proceedings of the 2017 Conference on Empirical Methods in
  Natural Language Processing}, 1957--1967. Copenhagen, Denmark: Association
  for Computational Linguistics.

\bibitem[{Bergsma and Lin(2006)}]{bergsma-lin-2006-bootstrapping}
Bergsma, S.; and Lin, D. 2006.
\newblock Bootstrapping Path-Based Pronoun Resolution.
\newblock In \emph{Proceedings of the 21st International Conference on
  Computational Linguistics and 44th Annual Meeting of the Association for
  Computational Linguistics}, 33--40. Sydney, Australia: Association for
  Computational Linguistics.

\bibitem[{Chen et~al.(2020)Chen, Lin, Li, Li, Zhou, and
  Sun}]{Chen_Lin_Li_Li_Zhou_Sun_2020}
Chen, D.; Lin, Y.; Li, W.; Li, P.; Zhou, J.; and Sun, X. 2020.
\newblock Measuring and Relieving the Over-Smoothing Problem for Graph Neural
  Networks from the Topological View.
\newblock \emph{Proceedings of the AAAI Conference on Artificial Intelligence},
  34(04): 3438--3445.

\bibitem[{Chomsky(1988)}]{chomsky_language_1988}
Chomsky, N. 1988.
\newblock \emph{Language and {Problems} of {Knowledge}: {The} {Managua}
  {Lectures}}.
\newblock Cambridge, MA: MIT Press.

\bibitem[{Clark and Manning(2016)}]{clark-manning-2016-improving}
Clark, K.; and Manning, C.~D. 2016.
\newblock Improving Coreference Resolution by Learning Entity-Level Distributed
  Representations.
\newblock In \emph{Proceedings of the 54th Annual Meeting of the Association
  for Computational Linguistics (Volume 1: Long Papers)}, 643--653. Berlin,
  Germany: Association for Computational Linguistics.

\bibitem[{Devlin et~al.(2019)Devlin, Chang, Lee, and
  Toutanova}]{devlin-etal-2019-bert}
Devlin, J.; Chang, M.-W.; Lee, K.; and Toutanova, K. 2019.
\newblock {BERT}: Pre-training of Deep Bidirectional Transformers for Language
  Understanding.
\newblock In \emph{Proceedings of the 2019 Conference of the North {A}merican
  Chapter of the Association for Computational Linguistics: Human Language
  Technologies, Volume 1 (Long and Short Papers)}, 4171--4186.

\bibitem[{Dror et~al.(2018)Dror, Baumer, Shlomov, and
  Reichart}]{dror-etal-2018-hitchhikers}
Dror, R.; Baumer, G.; Shlomov, S.; and Reichart, R. 2018.
\newblock The Hitchhiker{'}s Guide to Testing Statistical Significance in
  Natural Language Processing.
\newblock In \emph{Proceedings of the 56th Annual Meeting of the Association
  for Computational Linguistics (Volume 1: Long Papers)}, 1383--1392.
  Melbourne, Australia: Association for Computational Linguistics.

\bibitem[{Ge, Hale, and Charniak(1998)}]{Ge98astatistical}
Ge, N.; Hale, J.; and Charniak, E. 1998.
\newblock A Statistical Approach to Anaphora Resolution.
\newblock In \emph{In Proceedings of the Sixth Workshop on Very Large Corpora},
  161--170.

\bibitem[{Hobbs(1978)}]{hobbs-resolving-1987}
Hobbs, J. 1978.
\newblock Resolving pronoun references.
\newblock \emph{Lingua 44}, 311--338.

\bibitem[{Jiang and Cohn(2021)}]{jiang-cohn-2021-incorporating}
Jiang, F.; and Cohn, T. 2021.
\newblock Incorporating Syntax and Semantics in Coreference Resolution with
  Heterogeneous Graph Attention Network.
\newblock In \emph{Proceedings of the 2021 Conference of the North American
  Chapter of the Association for Computational Linguistics: Human Language
  Technologies}, 1584--1591. Online: Association for Computational Linguistics.

\bibitem[{Joshi et~al.(2020)Joshi, Chen, Liu, Weld, Zettlemoyer, and
  Levy}]{joshi-etal-2020-spanbert}
Joshi, M.; Chen, D.; Liu, Y.; Weld, D.~S.; Zettlemoyer, L.; and Levy, O. 2020.
\newblock {S}pan{BERT}: Improving Pre-training by Representing and Predicting
  Spans.
\newblock \emph{Transactions of the Association for Computational Linguistics},
  8: 64--77.

\bibitem[{Joshi et~al.(2019)Joshi, Levy, Zettlemoyer, and
  Weld}]{joshi-etal-2019-bert}
Joshi, M.; Levy, O.; Zettlemoyer, L.; and Weld, D. 2019.
\newblock {BERT} for Coreference Resolution: Baselines and Analysis.
\newblock In \emph{Proceedings of the 2019 Conference on Empirical Methods in
  Natural Language Processing and the 9th International Joint Conference on
  Natural Language Processing (EMNLP-IJCNLP)}, 5803--5808.

\bibitem[{Kingma and Ba(2015)}]{DBLP:journals/corr/KingmaB14}
Kingma, D.~P.; and Ba, J. 2015.
\newblock Adam: {A} Method for Stochastic Optimization.
\newblock In Bengio, Y.; and LeCun, Y., eds., \emph{3rd International
  Conference on Learning Representations, {ICLR} 2015}.

\bibitem[{Kong and Jian(2019)}]{ijcai2019-700}
Kong, F.; and Jian, F. 2019.
\newblock Incorporating Structural Information for Better Coreference
  Resolution.
\newblock In \emph{Proceedings of the Twenty-Eighth International Joint
  Conference on Artificial Intelligence, {IJCAI-19}}, 5039--5045. International
  Joint Conferences on Artificial Intelligence Organization.

\bibitem[{Kong and Zhou(2011)}]{kong-zhou-2011-combining}
Kong, F.; and Zhou, G. 2011.
\newblock Combining Dependency and Constituent-based Syntactic Information for
  Anaphoricity Determination in Coreference Resolution.
\newblock In \emph{Proceedings of the 25th Pacific Asia Conference on Language,
  Information and Computation}, 410--419. Singapore: Institute of Digital
  Enhancement of Cognitive Processing, Waseda University.

\bibitem[{Kong et~al.(2010)Kong, Zhou, Qian, and
  Zhu}]{kong-etal-2010-dependency}
Kong, F.; Zhou, G.; Qian, L.; and Zhu, Q. 2010.
\newblock Dependency-driven Anaphoricity Determination for Coreference
  Resolution.
\newblock In \emph{Proceedings of the 23rd International Conference on
  Computational Linguistics (Coling 2010)}, 599--607. Beijing, China: Coling
  2010 Organizing Committee.

\bibitem[{Lee et~al.(2017)Lee, He, Lewis, and Zettlemoyer}]{lee-etal-2017-end}
Lee, K.; He, L.; Lewis, M.; and Zettlemoyer, L. 2017.
\newblock End-to-end Neural Coreference Resolution.
\newblock In \emph{Proceedings of the 2017 Conference on Empirical Methods in
  Natural Language Processing}, 188--197. Association for Computational
  Linguistics.

\bibitem[{Lee, He, and Zettlemoyer(2018)}]{lee-etal-2018-higher}
Lee, K.; He, L.; and Zettlemoyer, L. 2018.
\newblock Higher-Order Coreference Resolution with Coarse-to-Fine Inference.
\newblock In \emph{Proceedings of the 2018 Conference of the North {A}merican
  Chapter of the Association for Computational Linguistics: Human Language
  Technologies, Volume 2 (Short Papers)}, 687--692.

\bibitem[{Lu and Ng(2020)}]{lu-ng-2020-conundrums}
Lu, J.; and Ng, V. 2020.
\newblock Conundrums in Entity Coreference Resolution: Making Sense of the
  State of the Art.
\newblock In \emph{Proceedings of the 2020 Conference on Empirical Methods in
  Natural Language Processing (EMNLP)}, 6620--6631. Online: Association for
  Computational Linguistics.

\bibitem[{Marcheggiani and Titov(2017)}]{marcheggiani-titov-2017-encoding}
Marcheggiani, D.; and Titov, I. 2017.
\newblock Encoding Sentences with Graph Convolutional Networks for Semantic
  Role Labeling.
\newblock In \emph{Proceedings of the 2017 Conference on Empirical Methods in
  Natural Language Processing}, 1506--1515. Copenhagen, Denmark: Association
  for Computational Linguistics.

\bibitem[{Marcheggiani and Titov(2020)}]{marcheggiani-titov-2020-graph}
Marcheggiani, D.; and Titov, I. 2020.
\newblock Graph Convolutions over Constituent Trees for Syntax-Aware Semantic
  Role Labeling.
\newblock In \emph{Proceedings of the 2020 Conference on Empirical Methods in
  Natural Language Processing (EMNLP)}, 3915--3928. Online: Association for
  Computational Linguistics.

\bibitem[{Ng(2010)}]{ng-2010-supervised}
Ng, V. 2010.
\newblock Supervised Noun Phrase Coreference Research: The First Fifteen Years.
\newblock In \emph{Proceedings of the 48th Annual Meeting of the Association
  for Computational Linguistics}, 1396--1411. Uppsala, Sweden: Association for
  Computational Linguistics.

\bibitem[{Peters et~al.(2018)Peters, Neumann, Iyyer, Gardner, Clark, Lee, and
  Zettlemoyer}]{peters-etal-2018-deep}
Peters, M.; Neumann, M.; Iyyer, M.; Gardner, M.; Clark, C.; Lee, K.; and
  Zettlemoyer, L. 2018.
\newblock Deep Contextualized Word Representations.
\newblock In \emph{Proceedings of the 2018 Conference of the North {A}merican
  Chapter of the Association for Computational Linguistics: Human Language
  Technologies, Volume 1 (Long Papers)}, 2227--2237.

\bibitem[{Pradhan et~al.(2012)Pradhan, Moschitti, Xue, Uryupina, and
  Zhang}]{pradhan-etal-2012-conll}
Pradhan, S.; Moschitti, A.; Xue, N.; Uryupina, O.; and Zhang, Y. 2012.
\newblock {C}o{NLL}-2012 Shared Task: Modeling Multilingual Unrestricted
  Coreference in {O}nto{N}otes.
\newblock In \emph{Joint Conference on {EMNLP} and {C}o{NLL} - Shared Task},
  1--40.

\bibitem[{Schlichtkrull et~al.(2017)Schlichtkrull, Kipf, Bloem, van~den Berg,
  Titov, and Welling}]{schlichtkrull2017modeling}
Schlichtkrull, M.; Kipf, T.~N.; Bloem, P.; van~den Berg, R.; Titov, I.; and
  Welling, M. 2017.
\newblock Modeling Relational Data with Graph Convolutional Networks.
\newblock arXiv:1703.06103.

\bibitem[{Socher et~al.(2013)Socher, Perelygin, Wu, Chuang, Manning, Ng, and
  Potts}]{socher-etal-2013-recursive}
Socher, R.; Perelygin, A.; Wu, J.; Chuang, J.; Manning, C.~D.; Ng, A.; and
  Potts, C. 2013.
\newblock Recursive Deep Models for Semantic Compositionality Over a Sentiment
  Treebank.
\newblock In \emph{Proceedings of the 2013 Conference on Empirical Methods in
  Natural Language Processing}, 1631--1642. Seattle, Washington, USA:
  Association for Computational Linguistics.

\bibitem[{Stoyanov et~al.(2009)Stoyanov, Gilbert, Cardie, and
  Riloff}]{stoyanov-etal-2009-conundrums}
Stoyanov, V.; Gilbert, N.; Cardie, C.; and Riloff, E. 2009.
\newblock Conundrums in Noun Phrase Coreference Resolution: Making Sense of the
  State-of-the-Art.
\newblock In \emph{Proceedings of the Joint Conference of the 47th Annual
  Meeting of the {ACL} and the 4th International Joint Conference on Natural
  Language Processing of the {AFNLP}}, 656--664. Suntec, Singapore: Association
  for Computational Linguistics.

\bibitem[{Tai, Socher, and Manning(2015)}]{tai-etal-2015-improved}
Tai, K.~S.; Socher, R.; and Manning, C.~D. 2015.
\newblock Improved Semantic Representations From Tree-Structured Long
  Short-Term Memory Networks.
\newblock In \emph{Proceedings of the 53rd Annual Meeting of the Association
  for Computational Linguistics and the 7th International Joint Conference on
  Natural Language Processing (Volume 1: Long Papers)}, 1556--1566. Beijing,
  China: Association for Computational Linguistics.

\bibitem[{Trieu et~al.(2019)Trieu, Duong~Nguyen, Nguyen, Miwa, Takamura, and
  Ananiadou}]{trieu-etal-2019-coreference}
Trieu, H.-L.; Duong~Nguyen, A.-K.; Nguyen, N.; Miwa, M.; Takamura, H.; and
  Ananiadou, S. 2019.
\newblock Coreference Resolution in Full Text Articles with {BERT} and
  Syntax-based Mention Filtering.
\newblock In \emph{Proceedings of The 5th Workshop on BioNLP Open Shared
  Tasks}, 196--205. Hong Kong, China: Association for Computational
  Linguistics.

\bibitem[{Veličković et~al.(2018)Veličković, Cucurull, Casanova, Romero,
  Liò, and Bengio}]{velikovi2018graph}
Veličković, P.; Cucurull, G.; Casanova, A.; Romero, A.; Liò, P.; and Bengio,
  Y. 2018.
\newblock Graph Attention Networks.
\newblock In \emph{International Conference on Learning Representations}.

\bibitem[{Wang et~al.(2020)Wang, Shen, Yang, Quan, and
  Wang}]{wang-etal-2020-relational}
Wang, K.; Shen, W.; Yang, Y.; Quan, X.; and Wang, R. 2020.
\newblock Relational Graph Attention Network for Aspect-based Sentiment
  Analysis.
\newblock In \emph{Proceedings of the 58th Annual Meeting of the Association
  for Computational Linguistics}, 3229--3238.

\bibitem[{Wang et~al.(2019)Wang, Yu, Zheng, Gan, Gai, Ye, Li, Zhou, Huang, Ma,
  Huang, Guo, Zhang, Lin, Zhao, Li, Smola, and
  Zhang}]{DBLP:journals/corr/abs-1909-01315}
Wang, M.; Yu, L.; Zheng, D.; Gan, Q.; Gai, Y.; Ye, Z.; Li, M.; Zhou, J.; Huang,
  Q.; Ma, C.; Huang, Z.; Guo, Q.; Zhang, H.; Lin, H.; Zhao, J.; Li, J.; Smola,
  A.~J.; and Zhang, Z. 2019.
\newblock Deep Graph Library: Towards Efficient and Scalable Deep Learning on
  Graphs.
\newblock \emph{CoRR}, abs/1909.01315.

\bibitem[{Wu et~al.(2020)Wu, Wang, Yuan, Wu, and Li}]{wu-etal-2020-corefqa}
Wu, W.; Wang, F.; Yuan, A.; Wu, F.; and Li, J. 2020.
\newblock {C}oref{QA}: Coreference Resolution as Query-based Span Prediction.
\newblock In \emph{Proceedings of the 58th Annual Meeting of the Association
  for Computational Linguistics}, 6953--6963.

\bibitem[{Wu and Gardner(2020)}]{wu2020understanding}
Wu, Z.; and Gardner, M. 2020.
\newblock Understanding Mention Detector-Linker Interaction for Neural
  Coreference Resolution.
\newblock arXiv:2009.09363.

\bibitem[{Xu et~al.(2015)Xu, Wang, Chen, and Li}]{xu2015empirical}
Xu, B.; Wang, N.; Chen, T.; and Li, M. 2015.
\newblock Empirical Evaluation of Rectified Activations in Convolutional
  Network.
\newblock arXiv:1505.00853.

\bibitem[{Xu and Choi(2020)}]{xu-choi-2020-revealing}
Xu, L.; and Choi, J.~D. 2020.
\newblock Revealing the Myth of Higher-Order Inference in Coreference
  Resolution.
\newblock In \emph{Proceedings of the 2020 Conference on Empirical Methods in
  Natural Language Processing (EMNLP)}, 8527--8533.

\bibitem[{Zhang, Zhou, and Li(2020)}]{zhang-etal-2020-fast}
Zhang, Y.; Zhou, H.; and Li, Z. 2020.
\newblock Fast and Accurate Neural {CRF} Constituency Parsing.
\newblock In \emph{Proceedings of IJCAI}, 4046--4053.

\end{thebibliography}

\end{document}